\documentclass[fleqn,10pt]{wlscirep}

\usepackage{soul}
\usepackage{color}
\usepackage{xcolor}
\usepackage{url}
\usepackage{listings}

\title{DLOPT: Deep Learning Optimization Library}

\author[1,*]{Andr\'es Camero}
\author[1]{Jamal Toutouh}
\author[1]{Enrique Alba}
\affil[1]{Universidad de M\'alaga, %
Departamento de Lenguajes y Ciencias de la Computaci\'on, %
M\'alaga, Espa\~na}
\affil[*]{andrescamero@uma.es}

\begin{abstract}
Deep learning hyper-parameter optimization is a tough task. Finding an appropriate network configuration is a key to success, however most of the times this labor is roughly done. In this work we introduce a novel library to  tackle this problem, the Deep Learning Optimization Library: DLOPT. We briefly describe its architecture and present a set of use examples. This is an open source project developed under the GNU~GPL~v3 license and it is freely available at \mbox{\url{https://github.com/acamero/dlopt}}.\\

\end{abstract}
\begin{document}

% ------------------------------------------------------------------------ %
\flushbottom
\maketitle
\thispagestyle{empty}

% ------------------------------------------------------------------------ %
\colorlet{punct}{red!60!black}
\definecolor{background}{HTML}{EEEEEE}
\definecolor{delim}{RGB}{20,105,176}
\colorlet{numb}{magenta!60!black}

\lstdefinelanguage{json}{
    basicstyle=\normalfont\ttfamily,
    numbers=left,
    numberstyle=\normalsize,
    stepnumber=1,
    numbersep=8pt,
    showstringspaces=false,
    breaklines=true,
    frame=lines
}
% ------------------------------------------------------------------------ %

% ------------------------------------------------------------------------ %
\section{Introduction}
\label{sec:introduction}

Hyper-parameter optimization is a challenging task, specially in the context of deep neural networks (DNN) and  deep learning (DL). Its main aim is to find/select the most suitable DNN configuration parameters for a specific  problem~\cite{Bergstra2011,Jozefowicz2015}. 
These parameters determine the activation functions, the number of hidden layers, the kernel size of a layer, etc.

The number of parameters and the large number of their possible values define a high-dimensional search space.  
Even though it is hard to find a ``good'' configuration, most DL studies in the literature rely on expert knowledge and on manual exploration strategies to address this problem~\cite{LeCun2015}.
Therefore, it seems mandatory to use an automatic intelligent tool to find an efficient hyper-parameter configuration (e.g., grid, evolutionary or random search) ~\cite{Albelwi2017,Domhan2015,Smithson2016}.

%\hl{TODO: Paragraph about automatic search algorithms}

In this context, \emph{evolutionary algorithms} (EAs)~\cite{back1996evolutionary} emerged as efficient stochastic methods to address hyper-parameter optimization problems. 
Indeed, these algorithms have been used in the literature to address this problem in shallower neural networks~\cite{alba2006metaheuristic,Yao1999}. 
However, these solutions are not appropriate in a DL context due to the high complexity of DNNs~\cite{Ojha2017}.

Nowadays, some new studies have analyzed the use of different variants of EAs to provide efficient DNN configurations, giving rise to \emph{deep neuroevolution}~\cite{Camero2018_LION,Miikkulainen2017,Morse2016,Such2017}. 
These studies have shown that EAs are adequate to deal with DNN hyper-parameter configuration problem and that they require less computational resources than traditional approaches (e.g. backpropagation, grid search).
Thus, these competitive results motivated us to develop a software library,
the \emph{Deep Learning Optimization Library (DLOPT)},
to ease the use of these promising algorithms to address hyper-parameter optimization without requiring a thorough knowledge in EAs.

In this report we briefly introduce DLOP, a library for deep learning hyper-parameter optimization. 
First, we outline its architecture. 
Then, Section~\ref{section:examples} presents some examples and summarizes their main results. 
Finally, Section~\ref{section:conclusions} presents the essential conclusions and formulates the principal lines for future work.

% ------------------------------------------------------------------------ %
\section{Core Architecture}\label{section:architecture}

The core of DLOPT is composed of three classes: \verb\ModelOptimization\, \verb\Problem\, and \verb\Solution\. This division aims to decouple the problem being solved (e.g. optimizing the architecture of a neural network) and the technique used to solve the problem (e.g. using a genetic algorithm to optimize the testing error of a neural network). 
At a glance, the \verb\ModelOptimization\ is an abstract class that is the basis for solving the hyper-parameter optimization problem. The \verb\Problem\ class is an abstract definition that acts as the basis for encoding, decoding and evaluating a solution. And finally, a \verb\Solution\ is a class that encapsulates an artificial neural network encoded by a specific problem, as well as the metrics (values of the fitness) calculated.

The library presents multiple implementations of the referred classes, as well as many utilities (e.g. \emph{random samplings}~\cite{Camero2018_NIPS,Camero2018_CAEPIA}, recurrent neural network builders and trainers, and data loader, among others), and we are still working on new state-of-the-art techniques to tackle the hyper-parametrization optimization problem.

The implementations (of the core classes) are divided into two major categories: architecture and weight. Both problems are related to each other, however we decided to split the code to ease the usability. The \emph{architecture} submodule intends to solve the problem of optimizing the architecture of a given network, while the training (i.e. the optimization of the weights) is done by commonly used techniques (e.g. backpropagation-based methods). On the other hand, the \emph{weight} submodule focuses on solving the optimization of the synaptic weights. 
The library also includes a submodule named \emph{tools}. This module offers a basic command line interface to use some functionalities. This interface can be broaden by extending the abstract class \verb\ActionBase\. %Examples of the use of this tools can be found in the \emph{examples} folder and in Section~\ref{section:examples}.

DLOPT is implemented on Python~3 and it relies on Keras (version~2.1) and Tensorflow (version~1.3)~\cite{abadi2016tensorflow}. The source code is available under the GNU~GPL~v3 license (\mbox{\url{https://github.com/acamero/dlopt}}).

% ------------------------------------------------------------------------ %
\section{DLOPT in Action}\label{section:examples}

This section aims at illustrating the usage and the performance of DLOPT. 
In this case we focused on the use of our library to deal with a specific type of DNNs, recurrent neural networks (RNN). 
Thus, we tackled a well-known test problem in the domain of DL by using RNNs, the \emph{sine wave}. 
Despite its simplicity, we selected sine wave because by adding this type of waves it is possible to approximate any periodic waveform~\cite{bracewell1986fourier}.

In order to ease the reproductivity of the presented experimentation the library includes an executable file to run these examples. This file is located in the following path: \texttt{examples/sin/run.sh} (see Listing~\ref{list:runsh}). 
Two different examples are implemented to be run: 
the \emph{MAE random sampling}~\cite{Camero2018_NIPS,Camero2018_CAEPIA} (Line~2) and the hyper-parameter optimization of a RNN by using the \emph{MAE random sampling} (Line~3), in order to introduce the main features of our library. 

\begin{lstlisting}[language=sh,numbers=left, frame=single,caption=Shell script implemented in \texttt{run.sh}., label=list:runsh]
#!/bin/bash
python ../../dlopt/tools/main.py --config mae-rand-samp-sin.json --verbose=1
python ../../dlopt/tools/main.py --config mae-optimization-sin.json --verbose=1
\end{lstlisting}

Listing~\ref{list:maeresult} summarizes the results provided by the MAE random sampling of a given architecture. In this case, the RNN applies a look back of two (Line~1) and its architecture is defined by one neuron in the input layer, two hidden layers with two neurons each, and one neuron output layer (see Line~2). 
The main results of the sampling are shown in lines 3 to 9. 

\begin{lstlisting}[language=json,numbers=left, frame=single,caption=MAE random sampling results of a given architecture., label=list:maeresult]
"look_back": 2,
"architecture": [1, 2, 2, 1],
"metrics": {
	"log_p": -29.906005813483684,
	"p": 1.0279848149134584e-13,
	"mean": 0.81072833698353519,
	"samples": [0.61528207663056766, ..., 0.72354892475102595],
	"std": 0.1104613706995357
}
\end{lstlisting}

Listing~\ref{list:earesult} shows a portion of the results provided by the evolutionary hyper-parameter optimization example. 
The optimization algorithm returns neural network model (configuration) that can be easily loaded by using Keras interface, and therefore, it can be used to address any DL problem. 

\begin{lstlisting}[language=json,numbers=left, frame=single,caption=Evolutionary hyper-parameter optimization results., label=list:earesult]
"fitness": { 'log_p': -12.215031852558125},
"layers": [1, 12, 13, 9, 10, 12, 6, 1],
"look_back": 17,
"config": [{
  "class_name": "LSTM",
  "config": {
	"activation": "tanh",
	"units": 12,
...  
\end{lstlisting}

In this case, the resulted RNN hyper-parameter configuration to address sine wave problem resulted with a fitness value of~-12.215 (Line~1). 
The parameters presented in Listing~\ref{list:earesult} define a network with a look back of 17 (Line~3) and an architecture with six hidden layers with 12, 13, 9, 10, 12, and 6 neurons, respectively (Line~2).
From Line~4 upwards, the output specifies the configuration of each hidden layer. In the example in Listing~\ref{list:earesult}, the first hidden layer includes 12 LSTM neurons (Line~5) that apply \emph{tanh} as activation function (Line~7).

% ------------------------------------------------------------------------ %
\section{Conclusions}\label{section:conclusions}

In this work we introduce DLOPT, a library for deep learning hyper-parameter optimization. We briefly present its core architecture and a set of use examples.

Despite being a work in progress, DLOPT is fully operational. Up to date, the library has been used solely for academic purposes, however we plan to transfer it to the industry. Most of the hyper-parameter optimization techniques included in DLOPT are originals and are publicly available, i.e. the source code and several scientific publications~\cite{Camero2018_NIPS,Camero2018_LION,Camero2018_CAEPIA}. Therefore, we consider that our library presents enough evidence to be considered as a useful and valid tool for hyper-parameter optimization. 

The design of DLOPT is intended to be flexible enough to be extended and adapted to a wide variety of hyper-parameter problems, including architecture and weight optimizations. Hence, as future work we propose to extend the library by adding new custom designed optimization algorithms and to add new problems.

% ------------------------------------------------------------------------ %
\bibliography{dlopt}

% ------------------------------------------------------------------------ %
\section*{Acknowledgements}

This research was partially funded by Ministerio de Economía, Industria y Competitividad, Gobierno de
España, and European Regional Development Fund grant numbers TIN2016-81766-REDT (http://cirti.es) and TIN2017-88213-R (http://6city.lcc.uma.es).

% ------------------------------------------------------------------------ %
\section*{Author contributions statement}

Conceptualization, A.C. and J.T.; Software, A.C.; Validation, A.C. and J.T.; Investigation, A.C. and J.T.; Resources, E.A.; Writing–Original Draft Preparation, A.C. and J.T.; Writing–Review \& Editing, A.C., J.T. and E.A.; Supervision, E.A.; Funding Acquisition, E.A.

\end{document}